%% file: main.tex
\documentclass[journal,12pt,onecolumn,draftclsnofoot,]{IEEEtran}

\usepackage{amsmath,amsfonts}
\usepackage{algorithmic}
\usepackage[linesnumbered,ruled,vlined]{algorithm2e}
\usepackage{array}
\usepackage{amssymb}
\usepackage{amsmath}
\usepackage{amsthm}
\usepackage{bbm}
\usepackage{textcomp}
\usepackage{stfloats}
\usepackage{bm}
\usepackage{url}
\usepackage{verbatim}
\usepackage{graphicx}
\usepackage{mathtools}
\usepackage{mwe}
\usepackage{multirow}
\usepackage{cite}
\usepackage{xcolor}
\usepackage{subfigure}
\newtheorem{theorem}{Theorem}
\newtheorem{assumption}{Assumption}

\usepackage{amssymb}

\begin{document}

\title{Communication Efficient and Privacy-Preserving Federated Learning Based on Evolution Strategies}

\author{Guangchen Lan

\thanks{.}
}

\markboth{Course Project of CS59200: Distributed Optimization for Machine Learning, December~2022}%
{Shell \MakeLowercase{\textit{et al.}}: A Sample Article Using IEEEtran.cls for IEEE Journals}

\IEEEpubid{0000--0000/00\$00.00~\copyright~2021 IEEE}

\maketitle

\begin{abstract}

Federated learning (FL) is an emerging paradigm for training deep neural networks (DNNs) in distributed manners. Current FL approaches all suffer from high communication overhead and information leakage. In this work, we present a federated learning algorithm based on evolution strategies (FedES), a zeroth-order training method. Instead of transmitting model parameters, FedES only communicates loss values, and thus has very low communication overhead. Moreover, a third party is unable to estimate gradients without knowing the pre-shared seed, which protects data privacy. Experimental results demonstrate FedES can achieve the above benefits while keeping convergence performance the same as that with back propagation methods.

\end{abstract}


\input{sections/Introduction}

\input{sections/Background_and_Problem_Statement}

\input{sections/Proposed_Algorithms}

\input{sections/Convergence_Analysis}

\input{sections/Experimental_Evaluations}

\section{Conclusions}

We have proposed a federated learning algorithm based on evolution strategies (FedES). Clients and the server only transmit loss values in each round, and thus it costs little in communication. On the other hand, a third party is unable to estimate gradients without knowing the pre-shared seed, which prevents information leakage. With experimental results, we showed that FedES can achieve the above benefits while keeping convergence performance the same as that with back propagation methods.

\input{sections/Appendix}

\bibliographystyle{IEEEtran}
\bibliography{ref.bib}

\end{document}

%% file: sections/Introduction.tex
\section{Introduction}

Federated learning (FL) \cite{li2020federated} offers a promising solution to the challenges posed by data centralization, where users communicate locally trained models rather than raw datasets. Recent work has expanded the scope of FL to federated reinforcement learning \cite{lan2023improved, khodadadian2022federated}. The main challenges of FL are the high communication overhead and information leakage during the training process\cite{kairouz2021advances}. In particular, the uplink transmission is usually considered as the bottleneck \cite{speedtest}, while the downlink transmission can be done through broadcast \cite{kairouz2021advances}. On the other hand, most FL training algorithms rely on back propagation. However, back propagation-based FL algorithms are not applicable when gradient information is not available \cite{fang2022communication}.

To reduce communication overhead in FL, FedAvg \cite{mcmahan2017communication} sets each client to update models locally with certain steps, and thus requires fewer communication rounds. Using a regularization term, FedProx \cite{li2020fedprox} extends FedAvg for non-iid data settings with requirements of data similarity. \cite{lan2023} further uses tensor decomposition methods to compress deep learning models. But these local update methods have to face a client inconsistency problem \cite{wang2020fednova}, which significantly influences the convergence performance.

To prevent information leakage during communication, a differential privacy method is adopted in \cite{wei2020differential}. In \cite{truex2019hybrid}, differential privacy is combined with secure multiparty computation to protect privacy. However, the convergence performance has to suffer the impact of added random noise, especially for high privacy requirements.

To tackle black-box objectives, distributed zeroth-order optimization algorithms are recently proposed in \cite{fang2022communication, hajinezhad2019zone, tang2020distributed, yi2022zeroth}. \cite{hajinezhad2019zone} proposes the ZONE algorithm for non-convex objectives based on primal-dual methods. \cite{tang2020distributed} then adopts a gradient tracking technique with faster convergence rates. \cite{yi2022zeroth} further achieves a linear speedup convergence rate under Polyak-Łojasiewicz conditions. The local update method in FedAvg is then combined with zeroth-order methods for communication efficiency in \cite{fang2022communication}, while it requires high similarity over data sets. However, in previous works, a large number of model parameters or gradients are transmitted in each round and thus it suffers from high communication overhead. The privacy advantage of zeroth-order optimization is not exploited either. On the other hand, natural gradient descent has not been studied in previous zeroth-order optimization methods, which has the optimal performance for neural networks \cite{amari1998natural}.

Evolution strategy (ES) \cite{wierstra2014natural} estimates a descent direction (expected as natural gradients) for arbitrary black-box functions, and has been recently used in reinforcement learning \cite{salimans2017evolution, khadka2018evolution} and computation graphs \cite{vicol2021unbiased}. The previous works indicate that ES has a genetic parallel behavior, but it has not been exploited in federated neural network training.

In this paper, we propose a federated learning algorithm based on evolution strategies (FedES) with the following benefits:
\begin{itemize}
\item \textbf{Communication overhead}. The number of transmitted scalars from each client is equal to the number of mini-batches, which can be extremely smaller than the size of model parameters in conventional FL.
\item \textbf{Heterogeneity}. FedES keeps the same performance in non-iid and inconsistency settings.
\item \textbf{Privacy}. Without knowing the common random seed, a third-party attacker gets no information from the communication process.
\item \textbf{Zeroth-order Optimization}. As no back propagation is required, FedES is suitable for black-box objectives.
\end{itemize}

The remainder of this paper is organized as follows. Section II introduces some background and problem settings. In Section III, we propose our federated learning algorithms based on evolution strategies. In Section IV, we analyze the convergence rate of FedES. Section V shows the experimental results and we conclude this paper in Section VI.

%% file: sections/Background_and_Problem_Statement.tex
\section{Backgroud and Problem Statement}

\textbf{Notations}: Scalars, vectors, and matrices are denoted by lowercase, boldface lowercase, and boldface uppercase letters, respectively, e.g., $x \in \mathbb{R}$, $\bm{x} \in\mathbb{R}^{N}$, $\bm{X} \in\mathbb{R}^{N_1 \times N_2}$.

\subsection{Evolution Strategy}

Evolution Strategy (ES) is a family of methods to estimate a natural gradient for black-box functions \cite{nesterov2017random}. To update a DNN model $\bm{w} \in\mathbb{R}^{N}$, natural gradients can be estimated based on $n$ data samples $\{\xi^{i}\}_{i=1}^{n}$ as follows
\begin{align}
\label{ng_loss}
l^{i} &= f(\bm{w} + \bm{\epsilon}^{i}; \xi^{i}), \\
\widetilde{\bm{g}} &= \frac{1}{n\sigma^{2}} \sum_{i=1}^{n} l^{i} \bm{\epsilon}^{i},
\label{ng_estimate}
\end{align}
where $f(\cdot)$ is a loss function, and $\bm{\epsilon}^{i}$ is a perturbation that has the same size as $\bm{w}$ with $N$ i.i.d. Gaussian ${\mathcal N}(0, \sigma^2)$ samples. It is seen from \eqref{ng_loss}-\eqref{ng_estimate} that the natural gradient $\widetilde{\bm{g}}$ is a linear combination of the perturbations to the model $\bm{w}$, with the weights being the losses of the perturbed networks on the training data.

The estimate of $\widetilde{\bm{g}}$ in \eqref{ng_loss}-\eqref{ng_estimate} has a high variance, and thus antithetic sampling (AS) \cite{owen2013monte} can be used for variance reduction. AS perturbs the parameters twice in opposite directions using the same perturbation $\bm{\epsilon}^{i}$, and computes the loss as
\begin{align}
\label{ng_as_loss}
l^{i} &= \frac{1}{2} \big( f(\bm{w} + \bm{\epsilon}^{i};\xi^{i}) - f(\bm{w} - \bm{\epsilon}^{i};\xi^{i}) \big), \\
 \widetilde{\bm{g}} &= \frac{1}{n\sigma^{2}} \sum_{i=1}^{n} l^{i} \bm{\epsilon}^{i}.
\label{ng_as_estimate}
\end{align}

The model parameters are then updated as 
\begin{align}
\label{ng_global_update}
\bm{w} \leftarrow \bm{w} - \alpha \widetilde{\bm{g}},
\end{align}
where $\alpha$ is the learning rate parameter.

\subsection{Problem Formulation}

We consider $K$ clients that cooperatively train a DNN model $\bm{w}$ with $N$ parameters. Assume client $k$ has $n_{k}$ data samples $\{\xi^{i}\}_{i=1}^{n_{k}}$. The goal is to minimize the global training loss, which is evaluated as follows
\begin{align}
\label{objective}
\min_{\bm{w} \in\mathbb{R}^{N}} \mathcal{L}(\bm{w}) &\coloneqq \sum_{k=1}^{K} \rho_{k} \mathcal{L}_{k}(\bm{w}),
\end{align}
where $\mathcal{L}_{k}(\bm{w}) = \frac{1}{n_k} \sum_{i=1}^{n_k} f(\bm{w};\xi^{i})$ is the averaged local loss based on the local data set from client $k$. The weight $\rho_k$ specifies the relative impact of client $k$, with one natural setting $\rho_k = \frac{n_k}{n}$, $k = 1,\ \cdots ,\ K$, where $n = \sum_{k=1}^{K} n_k$ is the total number of data samples \cite{li2020federated}.

%% file: sections/Proposed_Algorithms.tex
\section{Proposed Algorithms}

\begin{algorithm}
\caption{FedES}
\label{ng_fl_client}
\SetKwInput{KwServer}{Server executes}
 \SetKwInput{KwClient}{ClientUpdate$(k,\bm{w})$}
\KwServer{\\}

\begin{algorithmic}[1]
\STATE Initialize $\bm{w}$.
\STATE \textbf{for} $t=0,1,\cdots ,T$ \textbf{do}
\STATE ~~~~ \textbf{for} each client $k$ \textbf{in  parallel do}
\STATE ~~~~~~~~ $\{l^{b}_{k}\}_{b=1}^{B_{k}} \gets$ ClientUpdate$(k,\bm{w})$
\STATE ~~~~ \textbf{end}
\STATE ~~~~$\widetilde{\bm{g}} \gets \frac{1}{\sigma^{2}} \sum_{k=1}^{K} \frac{\rho_{k}}{B_{k}} \sum_{b=1}^{B_{k}} \bm{\epsilon}^{b}_{k} l^{b}_{k}$
\STATE ~~~~$\bm{w} \gets \bm{w} - \alpha \widetilde{\bm{g}}$
\STATE \textbf{end}
\end{algorithmic}

\KwClient{\\}
\begin{algorithmic}[1]
\STATE \textbf{for} $b=1,\cdots ,B_{k}$ \textbf{do}
\STATE ~~~~ $l_{k}^{b} \gets \frac{1}{2 n_B} \sum_{i=1}^{n_B} \big(f(\bm{w} + \bm{\epsilon}^{b}_{k};\xi^{b}_{i}) - f(\bm{w} - \bm{\epsilon}^{b}_{k};\xi^{b}_{i})\big)$
\STATE \textbf{end}
\STATE Transmit $\{l^{b}_{k}\}_{b=1}^{B_{k}}$ to server.
\end{algorithmic}
\end{algorithm}

The FedES algorithm is given in Algo. \ref{ng_fl_client}. Each client $k$ divides $n_{k}$ data samples into $B_{k}$ batches $\{\xi^{b}_{i}\}_{i=1}^{n_{B}}$, $b=1,\cdots ,B_{k}$, where $n_B = \frac{n_k}{B_k}$ is the common batch size. In the training process, the server pre-shares a common seed to all $K$ clients, which is used to generate a random seed$_k$ for client $k$ in each round. Each communication round $t$ consists of the following steps: 

\begin{enumerate}
\item The server broadcasts the current model parameters $\bm{w}$ to all $K$ clients.
\item Client $k$ generates a random seed$_k$ using the common seed, and then uses seed$_k$ to generate $B_{k}$ perturbations $\{\bm{\epsilon}_{k}^{b}\}_{b=1}^{B_{k}}$. Each perturbation contains $N$ i.i.d. Gaussian $\mathcal{N}(0, \sigma^2)$ samples. Client $k$ then performs the forward pass using parameters $\{\bm{w} + \bm{\epsilon}_{k}^{b}, \bm{w} - \bm{\epsilon}_{k}^{b}\}_{b=1}^{B_{k}}$ on its local training data set, and obtains the loss from all batches, $\{l_{k}^{b}\}_{b=1}^{B_{k}}$.
\item Client $k$ transmits $\{l_{k}^{b}\}_{b=1}^{B_{k}}$ to the server.
\item After receiving losses from all $K$ clients, $\{l_{k}^{b}\}_{b=1}^{B_{k}}$, $k=1,\cdots ,K$, the server generates $\{\bm{\epsilon}^{b}_{k}\}_{b=1}^{B_{k}}$, $k=1,\cdots ,K$, and estimates the natural gradient as follows
\begin{align}
\label{ng_algo}
\widetilde{\bm{g}}= \frac{1}{\sigma^{2}} \sum_{k=1}^{K} \frac{\rho_{k}}{B_{k}} \sum_{b=1}^{B_{k}} \bm{\epsilon}^{b}_{k} l^{b}_{k}.
\end{align}
The server then updates model parameters $\bm{w}$ according to \eqref{ng_global_update}.
\end{enumerate}

As perturbations $\{\bm{\epsilon}^{b}_{k}\}_{b=1}^{B_{k}}$, $k=1,\cdots ,K$ are generated from the pre-shared common seed, a third party is unable to know perturbation directions without knowing the seed, and then is unable to calculate $\widetilde{\bm{g}}$ or local gradient $\frac{1}{B_{k}} \sum_{b=1}^{B_{k}} \bm{\epsilon}^{b}_{k} l^{b}_{k}$.

\textbf{Elite Selection.} We can further reduce the communication overhead of FedES with elite selection. Instead of transmitting all loss values $\{l^{b}_{k}\}_{b=1}^{B_{k}}$ to the server in each round, client $k$ can select the $\beta B_{k}$ largest absolute values and only transmit selected loss values to the server, where $\beta$ is the elite rate. In an extreme case, client $k$ only transmits the largest loss value and $\beta B_{k} = 1$. We will show the performances with elite selection in Section V.

%% file: sections/Convergence_Analysis.tex
\section{Convergence Analysis}

In this section, we guarantee the convergence of Algorithm 1. Under Assumption \ref{taylor-assumption}, the convergence rate is given in Theorem \ref{theorem_convergence_rate}.

\begin{assumption}
\label{taylor-assumption}
Taylor's theorem applies to the gradients of the global loss function $\mathcal{L}$ as follows
\begin{align}
\nabla \mathcal{L}(\bm{w}) = \nabla \mathcal{L}(\bm{w}^{*}) + \nabla \big(\nabla \mathcal{L}(\bm{w})\big) (\bm{w} - \bm{w}^{*}) + \mathcal{O}\big((\bm{w} - \bm{w}^{*})^{2}\big),\ \forall \bm{w} \in\mathbb{R}^{N},
\end{align}
where $\mathcal{L}$ achieves the global minimum at $\bm{w}^{*}$.
\end{assumption}

\begin{theorem}
\label{theorem_convergence_rate}
The global loss function is up-bounded as follows
\begin{align}
\label{convergence_rate}
\mathbb{E}[\mathcal{L}(\bm{w}^{t}) - \mathcal{L}(\bm{w}^{*})] &\leq \mathcal{O}(\frac{1}{t}),
\end{align}
where $\bm{w}^{t}$ denotes model parameters in the $t$-th communication round.
\end{theorem}
The proof of Theorem \ref{theorem_convergence_rate} is given in the Appendix.

%% file: sections/Experimental_Evaluations.tex
\section{Experimental Evaluations}

\subsection{Experimental Setup}

\textbf{Data sets.} We use the following data sets:
\begin{itemize}
    \item MNIST data set \cite{lecun1998mnist} contains gray-scale images of handwritten digits, where each image has the size of $28 \times 28$. There are $60,000$ training images and $10,000$ testing images. Both are evenly split into $10$ classes.
    \item CIFAR-10 (not done yet) data set \cite{krizhevsky2009cifar} contains $60,000$ color images in $10$ classes, where each image has the size of $32\times 32\times 3$. There are $50,000$ training images and $10,000$ testing images. Both are evenly split into $10$ classes.
\end{itemize}

\textbf{Neural network settings.} There are two fully connected (FC) layers and one output layer, where each FC layer has a width $1024$, and the output layer has a width $10$. The first FC layer has a weight matrix size $784 \times 1024$ and a bias size $1024$. The second FC layer has a weight matrix size $1024 \times 1024$ and a bias size $1024$. The output layer has a weight matrix size $1024 \times 10$ and a bias size ${10}$. There are in total $N=1,863,690$ model parameters. The activation function is ReLU and the loss function is cross-entropy. The learning rate $\alpha$ is $0.01$.

\textbf{Performance metrics.} We consider the following performance metrics:
\begin{itemize}
\item Convergence: the loss value versus the communication round $t$ during the training process;
\item Test accuracy: the percentage of correctly estimated labels for samples in the testing data set;
\item Communication overhead: the number of parameters transmitted from each client.
\end{itemize}

\subsection{Numerical Results}

\begin{figure}
\centering
\subfigure[iid]{
\begin{minipage}[b]{1\linewidth}
\centering
\includegraphics[width=0.5\textwidth]{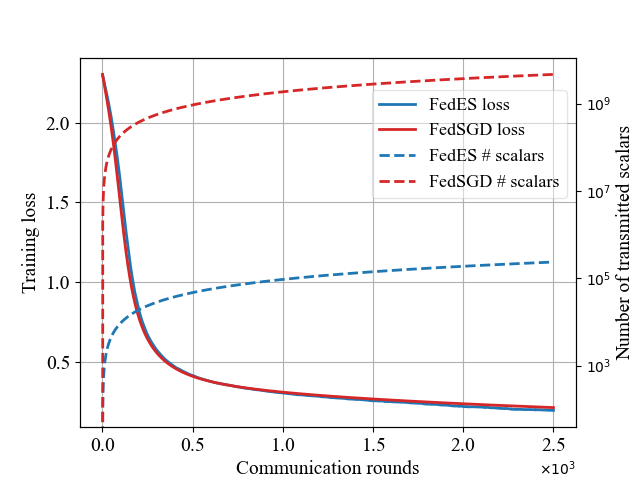}
\end{minipage}
}
\subfigure[non-iid]{
\begin{minipage}[b]{1\linewidth}
\centering
\includegraphics[width=0.5\textwidth]{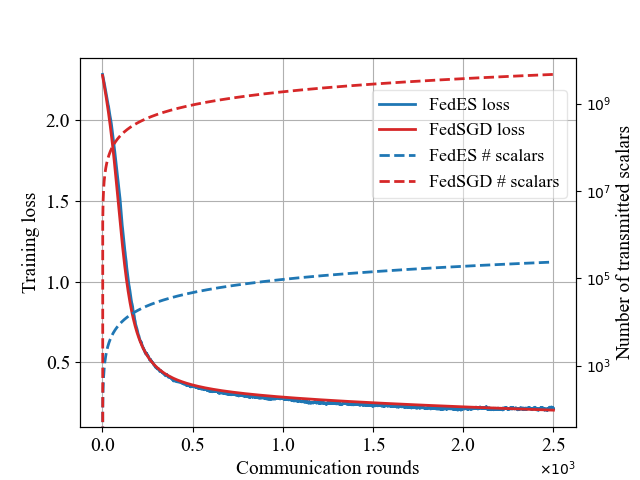}
\end{minipage}
}
\caption{Training loss and communication overhead of FedES and FedGD on the MNIST data set.}
\label{fedes-fedsgd}
\end{figure}

\textbf{Convergence.} As shown in Fig \ref{fedes-fedsgd}, we set $n_B = 64$ and compare FedES with the conventional FedGD (gradient descent) \cite{mcmahan2017communication}. The convergence performances have no significant difference between FedES and FedGD. At the same time, the number of transmitted data in FedES is about $2\times 10^{4}$ times smaller than that in FedGD.

\textbf{Trade-off between communication overhead and convergence performances.} The number of transmitted data from client $k$ equals the number of batches $B_k = \frac{n_k}{n_B}$. Thus, the smaller batch size $n_B$ makes communication overhead larger in each round, but gets estimates of natural gradients with lower variance as a return. The trade-off between communication overhead and convergence performance is shown in Table \ref{batch_table}.
\begin{table}
\centering
\caption{Test accuracies with different batch sizes on the MNIST data set.}
\label{batch_table}
\begin{tabular}{|c|c|c|c|}
\hline
\multirow{2}{*}{$n_B$} & \multirow{2}{*}{\# Data} & \multicolumn{2}{|c|}{Test accuracy} \\ \cline{3-4}
~ & ~ & i.i.d. & non-i.i.d. \\ \hline
$64$ & $94$ & $95.64\%$ & $95.58\%$ \\\hline
$256$ & $24$ & $94.15\%$ & $94.13\%$ \\\hline
$1024$ & $6$ & $93.76\%$ & $93.90\%$ \\\hline
\end{tabular}
\end{table}


\textbf{Limitation.} Experimental results are only given with toy examples for demonstration. The performance in large-scale systems can be further studied.

%% file: sections/Appendix.tex
\begin{appendices}

\section*{Appendix}

\subsection{Backgrounds of Natural Gradients}

In this subsection, we show the advantages of natural gradients in neural networks from previous work in \cite{amari1998natural}. Let $\mathbf{S} = \{\bm{w} \in\mathbb{R}^{N}\}$ be a parameter space where $\mathcal{L}(\bm{w})$ is defined. Proved in \cite{amari1998natural} and \cite{amari2012differential}, in multi-layer neural networks, the Riemannian structure of $\mathbf{S}$ is given by the Fisher information matrix $F(\bm{w}) \in \mathbb{R}^{N\times N}$. We assume $F(\bm{w})$ is invertible. A small increment $\mathrm{d}\bm{w}$ is then given by
\begin{align}
\label{increment}
\lVert \mathrm{d}\bm{w}\lVert^{2} = \sum_{i,j} F_{i,j}(\bm{w})\mathrm{d}w_{i}\mathrm{d}w_{j},
\end{align}
where $F_{i,j}(\bm{w})$ denotes the $(i,j)$-th entry of the matrix $F(\bm{w})$, and $w_i$, $w_j$ denotes the $i$-th, $j$-th element of $\bm{w}$, respectively. In Euclidean space, $F(\bm{w})$ is equal to the identity matrix. The steepest descent direction of $\mathcal{L}(\bm{w})$ is defined by
\begin{equation}
\begin{split}
\label{steepst}
&\mathop{\arg\min}\limits_{\mathrm{d}\bm{w}} \mathcal{L}(\bm{w} + \mathrm{d}\bm{w}), \\
&{\rm s.t.}\ \lVert \mathrm{d}\bm{w}\lVert^{2} = \varepsilon^{2},
\end{split}
\end{equation}
where $\varepsilon$ is a constant that can be arbitrarily small.

\begin{theorem}
The steepest descent direction of $\mathcal{L}(\bm{w})$ is given by
\begin{align}
\label{steepst-ng}
- \widetilde{\nabla} \mathcal{L}(\bm{w}) = - F^{-1}(\bm{w}) \cdot \nabla\mathcal{L}(\bm{w}),
\end{align}
where $\widetilde{\nabla} \mathcal{L}(\bm{w})$ denotes the natural gradient.
\end{theorem}

\textbf{Proof.} Let $\mathrm{d}\bm{w} = \varepsilon\bm{v}$, where $\lVert \bm{v}\lVert^{2} = \sum_{i,j} F_{i,j}(\bm{w})\mathrm{d}v_{i}\mathrm{d}v_{j} = 1$. The goal is transformed to minimize $\mathcal{L}(\bm{w} + \mathrm{d}\bm{w}) = \mathcal{L}(\bm{w}) + \varepsilon\nabla\mathcal{L}(\bm{w})^{T}\cdot\bm{v}$. Through the Lagrangian method, we have
\begin{align}
\label{Lagrangian}
\frac{\partial \big( \nabla \mathcal{L}(\bm{w})^{T}\cdot\bm{v} - \lambda\bm{v}^{T}F(\bm{w})\bm{v} \big)}{\partial v_{i}} = 0,\ i=1,\cdots ,N,
\end{align}
where $\lambda$ is the Lagrange multiplier. Then we have
\begin{align}
\label{Lagrangian-result}
\bm{v} = \frac{1}{2\lambda} F^{-1}(\bm{w})\cdot \nabla\mathcal{L}(\bm{w}).
\end{align}
The natural gradient with the steepest descent direction is then defined as
\begin{align}
\label{ng-define}
\widetilde{\nabla} \mathcal{L}(\bm{w}) = F^{-1}(\bm{w}) \cdot \nabla\mathcal{L}(\bm{w}).
\end{align}

\subsection{Proof of $\bf{Theorem\ \ref{theorem_convergence_rate}}$}

In Algorithm 1, after receiving loss values from clients, the server has
\begin{align}
\label{server-ng}
\widetilde{\bm{g}} = \frac{1}{\sigma^{2}} \sum_{k=1}^{K} \frac{\rho_{k}}{B_{k}} \sum_{b=1}^{B_{k}} \bm{\epsilon}^{b}_{k} l^{b}_{k}.
\end{align}
Taking $\rho_{k} = \frac{n_k}{n}$ and $B_{k} = \frac{n_k}{n_B}$, we have
\begin{align}
\label{server-rewrite}
\widetilde{\bm{g}} &= \frac{1}{\sigma^{2}n/n_B} \sum_{k=1}^{K}  \sum_{b=1}^{B_{k}} \bm{\epsilon}^{b}_{k} l^{b}_{k} \\
&= \frac{1}{\sigma^{2}n/n_B} \sum_{i=1}^{n/n_B} \bm{\epsilon}^{i} l^{i},
\label{folklore}
\end{align}
which performs overall data samples. Proved in \cite{rechenberg1989evolution, hansen2016cma}, $\widetilde{\bm{g}}$ is an unbiased estimator of $\widetilde{\nabla} \mathcal{L}(\bm{w})$, where $\widetilde{\nabla} \mathcal{L}(\bm{w}) = F^{-1}(\bm{w}) \cdot \nabla\mathcal{L}(\bm{w})$ is the natural gradient of $\mathcal{L}(\bm{w})$. The variance of the estimate is not easy to get for neural networks. But we can achieve an arbitrarily small variance as the number of searching directions $\frac{n}{n_B}$ is large enough. Thus, we consider the situation as $\widetilde{\bm{g}} = \widetilde{\mathcal{L}}(\bm{w})$ with large $\frac{n}{n_B}$.

\begin{theorem}
\label{var-theorem}
The expected squared error at the $t$-th iteration is given as
\begin{align}
\label{t-variance}
V^{t} = \mathbb{E}[(\bm{w}^{t} - \bm{w}^{*})(\bm{w}^{t} - \bm{w}^{*})^{T}] = \frac{1}{t}F^{-1}(\bm{w}^{*}) + \mathcal{O}(\frac{1}{t^2}),
\end{align}
when $\alpha$ in \eqref{ng_global_update} is chosen as $\frac{1}{t}$.
\end{theorem}

\textbf{Proof.} By Taylor's theorem, we have
\begin{align}
\label{gradient-taylor}
\nabla \mathcal{L}(\bm{w}^{t}) = \nabla \mathcal{L}(\bm{w}^{*}) + \nabla \big(\nabla \mathcal{L}(\bm{w}^{t})\big) (\bm{w}^{t} - \bm{w}^{*}) + \mathcal{O}\big((\bm{w}^{t} - \bm{w}^{*})^{2}\big).
\end{align}
At the $(t+1)$-th iteration, we have
\begin{align}
\label{t-ng}
\bm{w}^{t+1} = \bm{w}^{t} - \frac{1}{t}\widetilde{\nabla} \mathcal{L}(\bm{w}^{t}).
\end{align}
Subtracting $\bm{w}^{*}$ and taking the expected square operation on both sides, we have
\begin{align}
\label{t-ng-exp}
V^{t+1} = \mathbb{E}[\big( \bm{w}^{t} - \bm{w}^{*} + \frac{1}{t}\widetilde{\nabla} \mathcal{L}(\bm{w}^{t}) \big)\big( \bm{w}^{t} - \bm{w}^{*} + \frac{1}{t}\widetilde{\nabla} \mathcal{L}(\bm{w}^{t}) \big)^{T}].
\end{align}
Taking $\mathbb{E}[\nabla \mathcal{L}(\bm{w}^{*})] = \bm{0}$, $\mathbb{E}[\nabla \big(\nabla \mathcal{L}(\bm{w}^{*})\big)] = F(\bm{w}^{*})$, and $F^{-1}(\bm{w}^{t}) = F^{-1}(\bm{w}^{*}) + \mathcal{O}(\frac{1}{t})$, we achieve
\begin{align}
\label{t-ng-mse}
V^{t+1} = V^{t} - \frac{2}{t}V^{t} + \frac{1}{t^2}F^{-1}(\bm{w}^{*}) + \mathcal{O}(\frac{1}{t^3}).
\end{align}
Finally, the variance is given as
\begin{align}
\label{t-ng-var}
V^{t} = \frac{1}{t}F^{-1}(\bm{w}^{*}) + \mathcal{O}(\frac{1}{t^2}).
\end{align}

Now we show the analysis of the convergence rate based on Theorem \ref{var-theorem}. By Taylor's theorem and $\nabla \mathcal{L}(\bm{w}^{*}) = 0$, we have
\begin{equation}
\begin{split}
\label{k-loss}
\mathcal{L}(\bm{w}^{t}) - \mathcal{L}(\bm{w}^{*}) &= \frac{1}{2}(\bm{w}^{t} - \bm{w}^{*})^{T}H^{*}(\bm{w}^{t} - \bm{w}^{*}) + \nabla\mathcal{L}(\bm{w}^{*})^{T}(\bm{w}^{t} - \bm{w}^{*}) + \mathcal{O}\big((\bm{w}^{t} - \bm{w}^{*})^{3}\big)\\
&= \frac{1}{2}(\bm{w}^{t} - \bm{w}^{*})^{T}H^{*}(\bm{w}^{t} - \bm{w}^{*}) + \mathcal{O}\big((\bm{w}^{t} - \bm{w}^{*})^{3}\big),
\end{split}
\end{equation}
where $H^{*} \in\mathbb{R}^{N\times N}$ denotes the Hessian matrix at $\bm{w}^{*}$. Taking expectation operation on both sides, $H^{*}=F(\bm{w}^{*})$ \cite{amari1998natural}, and Theorem \ref{var-theorem}, we have
\begin{equation}
\begin{split}
\label{t-espect}
\mathbb{E}[\mathcal{L}(\bm{w}^{t})] - \mathcal{L}(\bm{w}^{*})
&= \frac{1}{2}\mathbb{E}[(\bm{w}^{t} - \bm{w}^{*})^{T}H^{*}(\bm{w}^{t} - \bm{w}^{*})] + \mathbb{E}[\mathcal{O}\big((\bm{w}^{t} - \bm{w}^{*})^{3}\big)] \\
&= \frac{1}{2}{\rm Tr}\Big( H^{*}\ \mathbb{E}[(\bm{w}^{t} - \bm{w}^{*})^{T}(\bm{w}^{t} - \bm{w}^{*})] \Big) + \mathbb{E}[\mathcal{O}\big((\bm{w}^{t} - \bm{w}^{*})^{3}\big)] \\
&= \frac{1}{2t}{\rm Tr}\Big( H^{*}F^{-1}(\bm{w}^{*}) \Big) + \mathbb{E}[\mathcal{O}\big((\bm{w}^{t} - \bm{w}^{*})^{3}\big)] \\
&= \frac{N}{2t} + \mathcal{O}(\frac{1}{t}),
\end{split}
\end{equation}
where ${\rm Tr}(\cdot)$ denotes the trace operation. As $\mathbb{E}[\mathcal{O}\big((\bm{w}^{t} - \bm{w}^{*})^{2}\big)] = \mathcal{O}(\frac{1}{t})$, the higher order term in the last second line should not converge slower in bounded regions.

\end{appendices}

%% file: main.bbl
\begin{thebibliography}{10}
\providecommand{\url}[1]{#1}
\csname url@samestyle\endcsname
\providecommand{\newblock}{\relax}
\providecommand{\bibinfo}[2]{#2}
\providecommand{\BIBentrySTDinterwordspacing}{\spaceskip=0pt\relax}
\providecommand{\BIBentryALTinterwordstretchfactor}{4}
\providecommand{\BIBentryALTinterwordspacing}{\spaceskip=\fontdimen2\font plus
\BIBentryALTinterwordstretchfactor\fontdimen3\font minus
  \fontdimen4\font\relax}
\providecommand{\BIBforeignlanguage}[2]{{%
\expandafter\ifx\csname l@#1\endcsname\relax
\typeout{** WARNING: IEEEtran.bst: No hyphenation pattern has been}%
\typeout{** loaded for the language `#1'. Using the pattern for}%
\typeout{** the default language instead.}%
\else
\language=\csname l@#1\endcsname
\fi
#2}}
\providecommand{\BIBdecl}{\relax}
\BIBdecl

\bibitem{li2020federated}
T.~Li, A.~K. Sahu, A.~Talwalkar, and V.~Smith, ``Federated learning:
  Challenges, methods, and future directions,'' \emph{IEEE Signal Processing
  Magazine}, vol.~37, no.~3, pp. 50--60, 2020.

\bibitem{lan2023improved}
G.~Lan, H.~Wang, J.~Anderson, C.~Brinton, and V.~Aggarwal, ``Improved
  communication efficiency in federated natural policy gradient via
  {ADMM}-based gradient updates,'' in \emph{Thirty-seventh Conference on Neural
  Information Processing Systems (NeurIPS)}, 2023.

\bibitem{khodadadian2022federated}
S.~Khodadadian, P.~Sharma, G.~Joshi, and S.~T. Maguluri, ``Federated
  reinforcement learning: Linear speedup under {Markovian} sampling,'' in
  \emph{International Conference on Machine Learning (ICML)}, 2022.

\bibitem{kairouz2021advances}
P.~Kairouz, H.~B. McMahan, B.~Avent, A.~Bellet, M.~Bennis, A.~N. Bhagoji,
  K.~Bonawitz, Z.~Charles, G.~Cormode, R.~Cummings \emph{et~al.}, ``Advances
  and open problems in federated learning,'' \emph{Foundations and
  Trends{\textregistered} in Machine Learning}, vol.~14, no. 1--2, pp. 1--210,
  2021.

\bibitem{speedtest}
speedtest.net, ``Speedtest {United States} market report.''\hskip 1em plus
  0.5em minus 0.4em\relax http://www.speedtest.net/reports/united-states/,
  2022.

\bibitem{fang2022communication}
W.~Fang, Z.~Yu, Y.~Jiang, Y.~Shi, C.~N. Jones, and Y.~Zhou,
  ``Communication-efficient stochastic zeroth-order optimization for federated
  learning,'' \emph{arXiv preprint arXiv:2201.09531}, 2022.

\bibitem{mcmahan2017communication}
B.~McMahan, E.~Moore, D.~Ramage, S.~Hampson, and B.~A.~y. Arcas,
  ``{Communication-Efficient Learning of Deep Networks from Decentralized
  Data},'' in \emph{Proceedings of the 20th International Conference on
  Artificial Intelligence and Statistics}, vol.~54, 20-22 Apr 2017, pp.
  1273--1282.

\bibitem{li2020fedprox}
T.~Li, A.~K. Sahu, M.~Zaheer, M.~Sanjabi, A.~Talwalkar, and V.~Smith,
  ``Federated optimization in heterogeneous networks,'' in \emph{Proceedings of
  Machine Learning and Systems}, vol.~2, 2020, pp. 429--450.

\bibitem{lan2023}
G.~Lan, X.-Y. Liu, Y.~Zhang, and X.~Wang, ``Communication-efficient federated
  learning for resource-constrained edge devices,'' \emph{IEEE Transactions on
  Machine Learning in Communications and Networking}, vol.~1, pp. 210--224,
  2023.

\bibitem{wang2020fednova}
J.~Wang, Q.~Liu, H.~Liang, G.~Joshi, and H.~V. Poor, ``Tackling the objective
  inconsistency problem in heterogeneous federated optimization,'' in
  \emph{Advances in Neural Information Processing Systems}, vol.~33, 2020, pp.
  7611--7623.

\bibitem{wei2020differential}
K.~Wei, J.~Li, M.~Ding, C.~Ma, H.~H. Yang, F.~Farokhi, S.~Jin, T.~Q.~S. Quek,
  and H.~V. Poor, ``Federated learning with differential privacy: Algorithms
  and performance analysis,'' \emph{IEEE Transactions on Information Forensics
  and Security}, vol.~15, pp. 3454--3469, 2020.

\bibitem{truex2019hybrid}
S.~Truex, N.~Baracaldo, A.~Anwar, T.~Steinke, H.~Ludwig, R.~Zhang, and Y.~Zhou,
  ``A hybrid approach to privacy-preserving federated learning,'' in
  \emph{Proceedings of the 12th ACM workshop on artificial intelligence and
  security}, 2019, pp. 1--11.

\bibitem{hajinezhad2019zone}
D.~Hajinezhad, M.~Hong, and A.~Garcia, ``Zone: Zeroth-order nonconvex
  multiagent optimization over networks,'' \emph{IEEE transactions on automatic
  control}, vol.~64, no.~10, pp. 3995--4010, 2019.

\bibitem{tang2020distributed}
Y.~Tang, J.~Zhang, and N.~Li, ``Distributed zero-order algorithms for nonconvex
  multiagent optimization,'' \emph{IEEE Transactions on Control of Network
  Systems}, vol.~8, no.~1, pp. 269--281, 2020.

\bibitem{yi2022zeroth}
X.~Yi, S.~Zhang, T.~Yang, and K.~H. Johansson, ``Zeroth-order algorithms for
  stochastic distributed nonconvex optimization,'' \emph{Automatica}, vol. 142,
  p. 110353, 2022.

\bibitem{amari1998natural}
S.~Amari, ``Natural gradient works efficiently in learning,'' \emph{Neural
  computation}, vol.~10, no.~2, pp. 251--276, 1998.

\bibitem{wierstra2014natural}
D.~Wierstra, T.~Schaul, T.~Glasmachers, Y.~Sun, J.~Peters, and J.~Schmidhuber,
  ``Natural evolution strategies,'' \emph{The Journal of Machine Learning
  Research}, vol.~15, no.~1, pp. 949--980, 2014.

\bibitem{salimans2017evolution}
T.~Salimans, J.~Ho, X.~Chen, S.~Sidor, and I.~Sutskever, ``Evolution strategies
  as a scalable alternative to reinforcement learning,'' \emph{arXiv preprint
  arXiv:1703.03864}, 2017.

\bibitem{khadka2018evolution}
S.~Khadka and K.~Tumer, ``Evolution-guided policy gradient in reinforcement
  learning,'' \emph{Advances in Neural Information Processing Systems},
  vol.~31, 2018.

\bibitem{vicol2021unbiased}
P.~Vicol, L.~Metz, and J.~Sohl-Dickstein, ``Unbiased gradient estimation in
  unrolled computation graphs with persistent evolution strategies,'' in
  \emph{International Conference on Machine Learning}.\hskip 1em plus 0.5em
  minus 0.4em\relax PMLR, 2021, pp. 10\,553--10\,563.

\bibitem{nesterov2017random}
Y.~Nesterov and V.~Spokoiny, ``Random gradient-free minimization of convex
  functions,'' \emph{Foundations of Computational Mathematics}, vol.~17, no.~2,
  pp. 527--566, 2017.

\bibitem{owen2013monte}
A.~B. Owen, \emph{Monte Carlo theory, methods and examples}.\hskip 1em plus
  0.5em minus 0.4em\relax Stanford, 2013.

\bibitem{lecun1998mnist}
Y.~LeCun, ``The {MNIST} database of handwritten digits,''
  \emph{http://yann.lecun.com/exdb/mnist/}, 1998.

\bibitem{krizhevsky2009cifar}
A.~Krizhevsky, V.~Nair, and G.~Hinton, ``The {CIFAR-10} dataset,''
  \emph{https://www.cs.toronto.edu/~kriz/cifar.html}, 2009.

\bibitem{amari2012differential}
S.~Amari, \emph{Differential-geometrical methods in statistics}.\hskip 1em plus
  0.5em minus 0.4em\relax Springer Science \& Business Media, 2012, vol.~28.

\bibitem{rechenberg1989evolution}
I.~Rechenberg, ``Evolution strategy: Nature’s way of optimization,'' in
  \emph{Optimization: Methods and applications, possibilities and
  limitations}.\hskip 1em plus 0.5em minus 0.4em\relax Springer, 1989, pp.
  106--126.

\bibitem{hansen2016cma}
N.~Hansen, ``The {CMA} evolution strategy: A tutorial,'' \emph{arXiv preprint
  arXiv:1604.00772}, 2016.

\end{thebibliography}
